\documentclass[pdflatex,sn-mathphys-num]{sn-jnl}

\usepackage{graphicx}%
\usepackage{multirow}%
\usepackage{amsmath,amssymb,amsfonts}%
\usepackage{amsthm}%
\usepackage{mathrsfs}%
\usepackage[title]{appendix}%
\usepackage{xcolor}%
\usepackage{textcomp}%
\usepackage{manyfoot}%
\usepackage{booktabs}%
\usepackage{algorithm}%
\usepackage{algorithmicx}%
\usepackage{algpseudocode}%
\usepackage{listings}%
\usepackage{makecell}
\usepackage{svg}

\theoremstyle{thmstyleone}%
\theoremstyle{thmstyletwo}%

\theoremstyle{thmstylethree}%

\begin{document}

\title[Hallucination Filtering in Radiology Vision-Language Models Using Discrete Semantic Entropy]{Hallucination Filtering in Radiology Vision-Language Models Using Discrete Semantic Entropy}


\makeatletter
\let\OldAffil\affil
\newcommand{\affilsize}{\fontsize{7}{7}\selectfont}
\renewcommand{\affil}[2][]{%
  \OldAffil[#1]{%
    \affilsize%
    \setlength{\parskip}{0pt}\setlength{\parindent}{0pt}%
    #2%
  }%
}
\makeatother

\author[1,2]{Patrick Wienholt}
\author[1,2]{Sophie Caselitz}
\author[1,2]{Robert Siepmann}
\author[2]{Philipp Bruners}
\author[3,4]{Keno Bressem}
\author[2]{Christiane Kuhl}
\author[5,6,7,8]{Jakob Nikolas Kather}
\author[1,2]{Sven Nebelung}
\author[1,2]{Daniel Truhn}

\affil[1]{Lab for Artificial Intelligence in Medicine, Department of Diagnostic and Interventional Radiology, University Hospital RWTH Aachen, Aachen, Germany}

\affil[2]{Department of Diagnostic and Interventional Radiology, University Hospital RWTH Aachen, Aachen, Germany}

\affil[3]{Department of Diagnostic and Interventional Radiology, Technical University of Munich, School of Medicine and Health, Klinikum rechts der Isar, TUM University Hospital, Munich, Germany}

\affil[4]{Department of Cardiovascular Radiology and Nuclear Medicine, Technical University of Munich, School of Medicine and Health, German Heart Center, TUM University Hospital, Munich, Germany}

\affil[5]{Else Kroener Fresenius Center for Digital Health, Faculty of Medicine and University Hospital Carl Gustav Carus, TUD Dresden University of Technology, Dresden, Germany}

\affil[6]{Department of Medicine I, Faculty of Medicine and University Hospital Carl Gustav Carus, TUD Dresden University of Technology, Dresden, Germany}

\affil[7]{Pathology \& Data Analytics, Leeds Institute of Medical Research at St James’s, University of Leeds, Leeds, United Kingdom}

\affil[8]{Medical Oncology, National Center for Tumor Diseases (NCT), University Hospital Heidelberg, Heidelberg, Germany}


\abstract{\unboldmath
\textbf{Purpose:}
To determine whether using discrete semantic entropy (DSE) to reject questions likely to generate hallucinations can improve the accuracy of black-box vision-language models (VLMs) in radiologic image based visual question answering (VQA).

\textbf{Methods:} 
This retrospective study evaluated DSE using two publicly available, de-identified datasets: the VQA-Med 2019 benchmark (500 images with clinical questions and short-text answers) and a diagnostic radiology dataset (206 cases: 60 computed tomography scans, 60 magnetic resonance images, 60 radiographs, 26 angiograms) with corresponding ground-truth diagnoses. GPT-4o and GPT-4.1 (Generative Pretrained Transformer; OpenAI) answered each question 15 times using a temperature of 1.0. Baseline accuracy was determined using low-temperature answers (temperature 0.1). Meaning-equivalent responses were grouped using bidirectional entailment checks, and DSE was computed from the relative frequencies of the resulting semantic clusters. Accuracy was recalculated after excluding questions with DSE $>$ 0.6 or $>$ 0.3. $p$-values and 95\% confidence intervals were obtained using bootstrap resampling and a Bonferroni-corrected threshold of $p < .004$ for statistical significance.

\textbf{Results:} 
Across 706 image–question pairs, baseline accuracy was 51.7\% for GPT-4o and 54.8\% for GPT-4.1. After filtering out high-entropy questions (DSE $>$ 0.3), accuracy on the remaining questions was 76.3\% (retained questions: 334/706) for GPT-4o and 63.8\% (retained questions: 499/706) for GPT-4.1 (both $p < .001$). Accuracy gains were observed across both datasets and largely remained statistically significant after Bonferroni correction.

\textbf{Conclusion:} 
DSE enables reliable hallucination detection in black-box VLMs by quantifying semantic inconsistency. This method significantly improves diagnostic answer accuracy and offers a filtering strategy for clinical VLM applications.

\textbf{Code:} \url{https://github.com/TruhnLab/VisionSemanticEntropy}
}


\maketitle

\section{Introduction}\label{sec1}
Growing examination volumes\cite{winder2021weS1} and complexity\cite{european2022roleS2} have drastically increased the workloads\cite{lantsman2022trendS3} of radiologists, with no  foreseeable resolution\cite{bruls2020workloadS4}. This challenge, compounded by a global radiologist shortage\cite{meng2023growingS5,yaghmai2024editorialS6}, raises concerns about burnout and potential diagnostic errors\cite{yaghmai2024editorialS6}. The early application of large language models\cite{bhayana2024chatbotsS7,han2024comparativeS8} in radiology focused on text-based tasks like suggesting diagnoses\cite{wang2025evaluationS9}, summarizing guidelines\cite{lim2024chatgptS10}, and extracting patient information\cite{wang2024entityS11}. However, given the inherently visual nature of radiology, the focus is rapidly shifting towards vision-language models\cite{DiagnosticAccS12} (VLMs) such as GPT-4o (Generative Pretrained Transformer; OpenAI), Gemini 2.5 Pro, and Claude 4, which integrate image analysis with linguistic capabilities. 

A critical challenge for the safe adoption of VLMs in radiology is their propensity to generate hallucinations\cite{Gunjal2024DetectionS13,farquhar2024detectingS14}: plausible-sounding outputs that are not grounded in visual evidence or clinical context. Unlike human experts, who express uncertainty, VLMs may present erroneous findings with high linguistic certainty, posing significant risks to diagnostic safety and clinician trust.
Reliance on human review for validation is not scalable for routine clinical use\cite{tanno2025collaborationS15}. Although uncertainty can be estimated using internal model mechanisms like token probabilities or activations\cite{Ji2023MAPS16,ZOU2023100003S17}, proprietary VLMs lack the transparency necessary for these approaches. Alternatively, auxiliary components like reward models can be used\cite{Gunjal2024DetectionS13}, but the need for additional training data or fine-tuning capabilities limits their feasibility for clinical practitioners. Consequently, hallucination-detection methods suitable for black-box scenarios are gaining traction. 

Among these methods, consistency-based techniques sample multiple outputs to identify inconsistencies, such as RadFlag for report generation\cite{pmlr-v259-zhang25cS18}, or rephrase input questions to check for response stability\cite{Khan2024BlackBoxS19}. A distinct, recent black-box approach specifically designed to detect hallucinations in text generation is discrete semantic entropy (DSE)\cite{farquhar2024detectingS14}. DSE quantifies uncertainty by measuring the semantic dispersion among multiple responses to the same prompt using high-temperature sampling, where the temperature parameter controls output randomness—higher values produce more variable responses, revealing potential model uncertainty. Responses are clustered by semantic meaning, and the entropy of this cluster distribution serves as an uncertainty score. Crucially, DSE does not require access to model internals or task-specific training data, and its assessments of semantic consistency are robust to simple linguistic paraphrasing\cite{farquhar2024detectingS14}. While DSE is validated for text, and related concepts like vision-amplified semantic entropy are emerging\cite{Liao2025VisionAplifiedS20}, the direct application and effectiveness of DSE for hallucination detection in clinical vision-language tasks, specifically radiologic interpretation, remain unexplored.

The objective of this study was to evaluate whether DSE could identify visual question answering tasks at high risk of hallucination and thereby improve the accuracy of accepted answers. We hypothesized that (i) applying DSE-based filtering would significantly increase the accuracy of accepted answers compared with baseline performance, (ii) that this improvement would be observed across multiple imaging modalities and question categories, and (iii) that stricter entropy thresholds would yield greater accuracy improvements at the expense of reduced coverage.

\section{Materials and methods}\label{sec2}
\subsection*{Study design}
\begin{figure}
\centering
\includegraphics[width=0.7\linewidth]{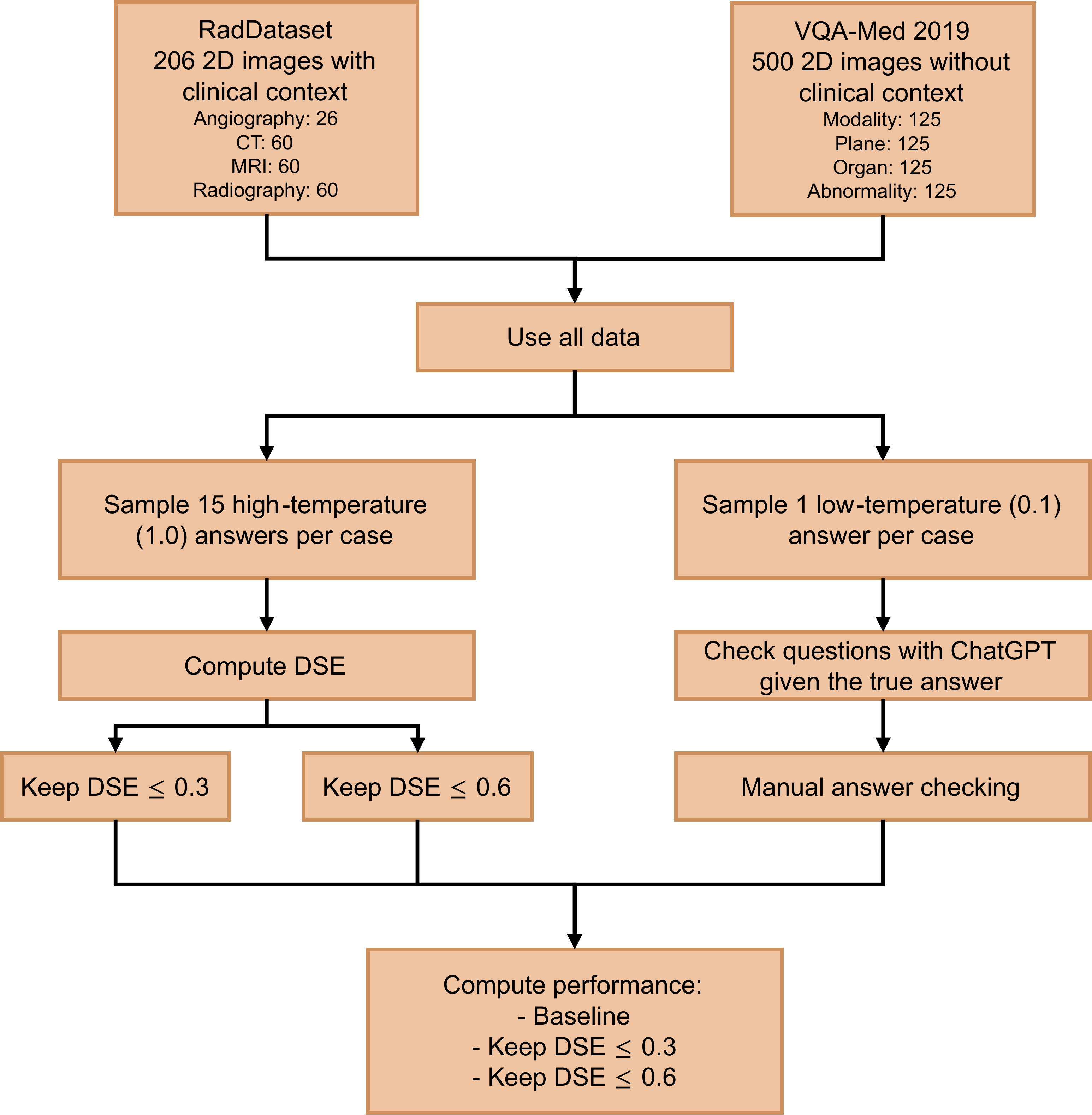}
\caption{Data flow and evaluation pipeline. All 206 clinical cases from the RadDataset and all 500 image–question pairs from VQA-Med 2019 were included. For each case, the model generated 15 high-temperature $(T = 1.0)$ responses to compute discrete semantic entropy (DSE), and questions were filtered at thresholds of $\leq 0.6$ and $\leq 0.3$. Baseline accuracy used a single low-temperature $(T = 0.1)$ response per case, which was compared with the reference answer using a GPT-based check and then verified manually. Performance was reported for the baseline and the two DSE-filtered settings.}
\label{fig:fig1}
\end{figure}

This retrospective study utilized data from two publicly available, de-identified datasets. Therefore, institutional review board approval and the requirement for informed consent were waived for this analysis. No studies were excluded. An overview of the study data flow and study design is depicted in Figure \ref{fig:fig1}. The first dataset was the VQA-Med 2019 test set, which comprised 500 radiological images, each paired with a clinically relevant question\cite{Abacha2019VQAMedS21}. These images originated from the National Library of Medicine’s MedPix\textsuperscript{®} teaching-file archive\cite{AnatomyTOOL_MedPixS22} and were de-identified according to NLM Web Policies\cite{NLM_WebPoliciesS23}. The questions were evenly distributed across four categories—modality, plane, organ, and abnormality—with 125 image–question pairs per category. All reference answers in the dataset were reviewed and validated by a radiologist\cite{Abacha2019VQAMedS21}. This study was conducted and reported in accordance with the Checklist for Artificial Intelligence in Medical Imaging (CLAIM).

The second dataset, referred to as RadDataset in the following, was a publicly available in-house dataset consisting of 206 selected de-identified clinical 2D images retrospectively collected from routine examinations as described by Huppertz et al.\cite{huppertz2025revolutionS24}. These included 60 computed tomography (CT) scans, 60 magnetic resonance (MR) images, 60 conventional radiographs, and 26 angiography studies. Each image was annotated with its clinical context and a corresponding diagnosis, which was confirmed by four radiologists in consensus\cite{huppertz2025revolutionS24}.

\subsection*{Assessment of semantic uncertainty}
We evaluated the semantic uncertainty of two OpenAI VLMs accessed via the Microsoft azure application programming interface (API): GPT‑4o (version: 2024-05-13) and GPT‑4.1 (version: 2025-04-14). For each model and image–question pair, we generated 15 independent responses at a temperature of 1.0. This high-temperature sampling procedure promoted output variability and was designed to reveal potential model uncertainty through inconsistent answers. Additionally, one response per pair was generated using a temperature of 0.1 to establish the baseline accuracy for each model. The accuracy of all VLM-generated answers was first reviewed by a medical student (S.C.) after thorough training by two senior board-certified radiologists (D.T., S.N.; 14 and 12 years of experience, respectively). All uncertain cases were escalated to these radiologists for adjudication. An answer was considered correct if its clinical meaning was semantically equivalent to the reference answer, even if its wording differed from the ground-truth text.

\begin{figure}[htbp]
  \centering
  \includegraphics[width=\textwidth]{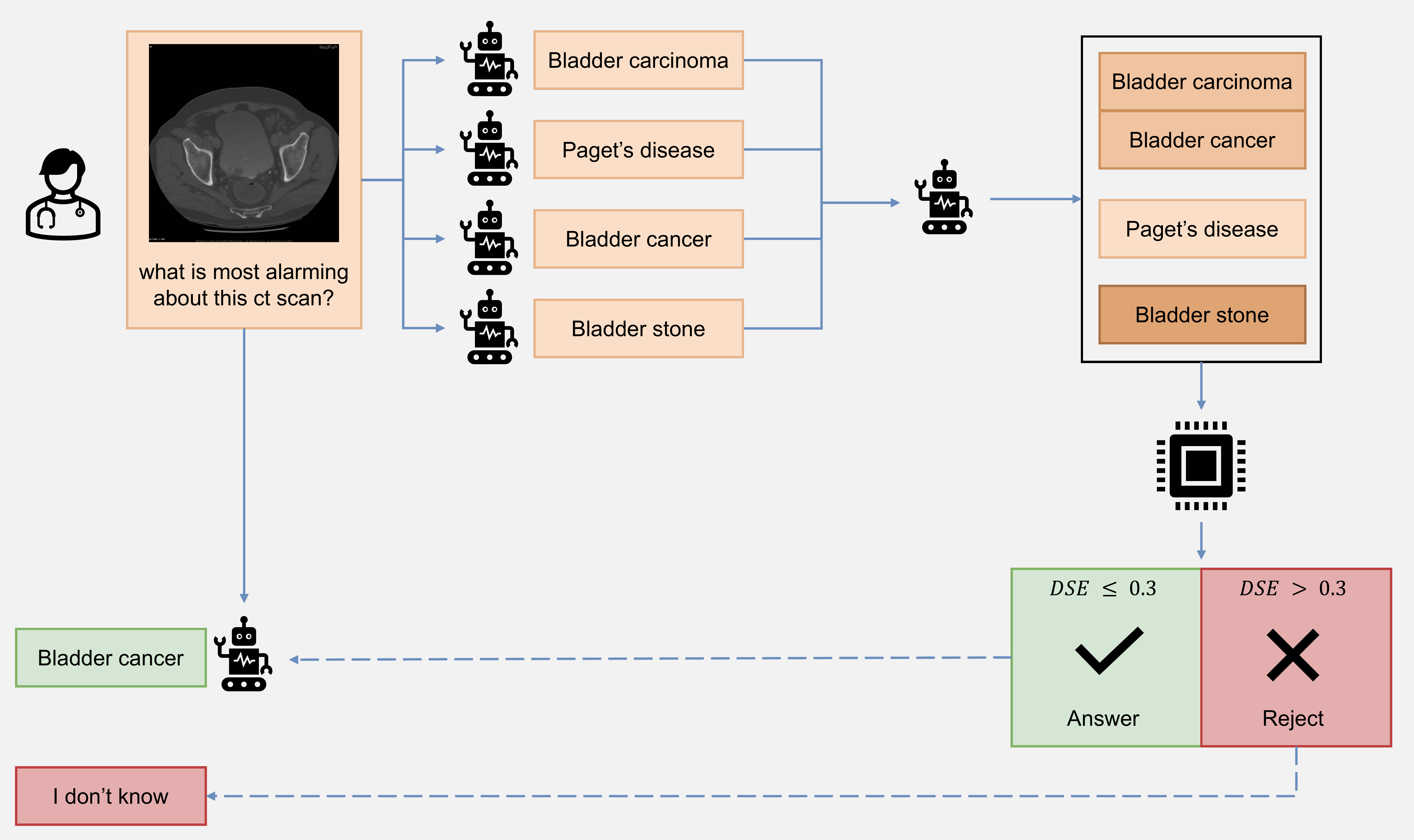}
  \caption{Schematic for discrete semantic entropy (DSE) implementation within clinical radiology workflows. A question asked by a clinician is posed 15 times to the vision-language model (VLM). The same model then clusters the answers based on semantic similarity. Based on the results of clustering, the DSE is calculated, and the question is answered or rejected by the VLM depending on the DSE threshold.}\label{fig:methodOverview}
\end{figure}

We then applied a semantic clustering procedure to group equivalent answers. The mutual entailment of each pair of generated responses was re-evaluated using the same VLM (either GPT‑4o or GPT‑4.1). Responses were assigned to the same semantic cluster only if mutual entailment was confirmed. This procedure ensured that semantically equivalent outputs were grouped together, thereby mitigating inflation of entropy caused by linguistic variation. All steps of the workflow illustrated in Figure \ref{fig:methodOverview} were automated.

\subsection*{Application of discrete semantic entropy}
The semantic clusters were used to compute the discrete semantic entropy (DSE) for each question. Let $C_i$ denote the $i$-th cluster and $|C_i |$  its number of assigned responses. The relative frequency of each cluster for a given question x was calculated as
\begin{equation}
    P(C_i|x) = \frac{|C_i|}{\sum_{C_k}|C_k|}.
\end{equation}
The entropy was then computed as
\begin{equation}
    \text{DSE}(x) = -\sum_{C_i}P(C_i|x)\log_{10}P(C_i|x).
\end{equation}
If all 15 responses were grouped into a single cluster, DSE was zero, indicating complete semantic consistency. The maximum theoretical entropy with 15 responses is $\log_{10}(15)\approx1.18$, which is reached when all answers are assigned to mutually different clusters, i.e., if each answer is its own cluster. DSE captures the semantic dispersion in the model’s responses and serves as a proxy for model uncertainty\cite{farquhar2024detectingS14}.  Importantly, DSE is applicable even in fully black-box settings, requiring only the model’s output without access to its internal parameters or probability distributions.

To assess whether DSE can serve as an indicator of hallucinations, we implemented a selective prediction strategy in which questions exceeding a predefined DSE threshold were discarded, and accuracy was recomputed on the remaining subset of questions. We performed this analysis using two threshold values—0.6 and 0.3—to evaluate how the choice of cutoff impacts accuracy. These thresholds were selected as they correspond to one half and one quarter of the maximum attainable DSE. For each of the two models, we applied both thresholds and evaluated them on each dataset individually as well as on the combined dataset, yielding a total of twelve comparisons.

\subsection*{Statistical analysis}
The statistical significance of accuracy improvements after entropy-based rejection was assessed using a two-sided bootstrap test with 100,000 iterations. Given that twelve comparisons were pre-specified, we applied Bonferroni correction to control the error rate, resulting in an adjusted significance threshold of $p< .05/12=.0042$. Confidence intervals (95\%) were estimated using the percentile bootstrap method. Our code is publicly available at \url{https://github.com/TruhnLab/VisionSemanticEntropy}.

\section{Results}
\subsection*{Baseline model accuracy}
Across the combined test set of 706 image–question pairs, baseline accuracy answering all questions (using sampling with temperature 0.1) was 51.7\% for GPT-4o and 54.8\% for GPT-4.1. Performance differed substantially between datasets. On the VQA-Med 2019 benchmark, GPT-4o and GPT-4.1 achieved 59.0\% and 63.2\% accuracy, respectively. Baseline accuracy on the RadDataset was low (34.0\% for GPT-4o and 34.5\% for GPT-4.1), highlighting the limitations of current general-purpose VLMs for medical image interpretation. Subcategory analysis revealed further differences in baseline performance (Table \ref{tab:subcategory-accuracy}). On VQA-Med 2019, both models showed strong baseline performance on questions about modality, imaging plane, and organs (accuracies ranging from 69.6\% to 83.2\%). However, the models struggled with abnormality-related questions, achieving baseline accuracies of only 13.6\% (GPT-4o) and 12.8\% (GPT-4.1). Baseline performance on the RadDataset also varied by modality, and both VLMs exhibited the highest accuracy on angiography studies and lowest accuracy on MR images. 

\subsection*{DSE application improves accuracy}
As Table \ref{tab:baseline-accuracy} shows, applying DSE-based filtering to reject questions with uncertain answers led to significant improvements in accuracy for the remaining questions. At a threshold of $\text{DSE} \leq 0.6$, GPT-4o accuracy on the combined dataset increased from 51.7\% to 62.9\% (retained 499 of 706 questions, $p < .001$), and GPT-4.1 accuracy improved from 54.8\% to 60.4\% (retained 626 of 706 questions, $p < .001$). Tightening the threshold to $\text{DSE} \leq 0.3$ further increased accuracy to 76.3\% for GPT-4o (retained questions: 334, $p < .001$) and 63.8\% for GPT-4.1 (retained questions: 499, $p < .001$). All improvements shown in Table \ref{tab:baseline-accuracy} were statistically significant after Bonferroni correction, except for GPT-4o on the RadDataset at $\text{DSE} \leq 0.3$, where the small number of remaining questions (n=29) limited statistical power.

\begin{table*}[t]
\centering
\caption{Baseline accuracy (Acc.) of the models, number of answered questions (\#), and accuracy improvement (\(\Delta\)) after filtering based on discrete semantic entropy (DSE) thresholds (\(\le 0.6\) and \(\le 0.3\)), with 95\% confidence intervals (CI). Bold p-values indicate statistical significance (\(p<.004\)).}
\label{tab:baseline-accuracy}
\resizebox{\textwidth}{!}{%
\begin{tabular}{llccccrrrrccrrrr}
\toprule
 &  & \multicolumn{2}{c}{All} & \multicolumn{6}{c}{DSE \(\le 0.6\)} & \multicolumn{6}{c}{DSE \(\le 0.3\)} \\
\cmidrule(lr){3-4} \cmidrule(lr){5-10} \cmidrule(lr){11-16}
Dataset & Model & \# & Acc.\ in \% & \# & Acc.\ in \% & \(\Delta\) in \% & \multicolumn{2}{c}{95\% CI in \%} & \(p\) & \# & Acc.\ in \% & \(\Delta\) in \% & \multicolumn{2}{c}{95\% CI in \%} & \(p\) \\
\midrule
\multirow{2}{*}{\shortstack{VQA\text{-}Med\\2019}} 
& GPT-4o   & 500 & 59.0 & 392 & 72.2 & \textbf{+13.2} & \textbf{+10.6} & \textbf{+16.0} & \(\mathbf{< .001}\) & 305 & 79.0 & \textbf{+20.0} & \textbf{+16.3} & \textbf{+23.9} & \(\mathbf{< .001}\) \\
& GPT-4.1  & 500 & 63.2 & 447 & 69.4 & \textbf{+6.2}  & \textbf{+4.4}  & \textbf{+8.1}  & \(\mathbf{< .001}\) & 373 & 76.1 & \textbf{+12.9} & \textbf{+10.1} & \textbf{+16.0} & \(\mathbf{< .001}\) \\
\midrule
\multirow{2}{*}{\shortstack{Rad\text{-}\\Dataset}} 
& GPT-4o   & 206 & 34.0 & 107 & 43.0 & \textbf{+9.0}  & \textbf{+2.9}  & \textbf{+15.4} & \(\mathbf{.003}\)   & 29  & 48.3 & +14.3 & -2.7 & 31.5 & .100 \\
& GPT-4.1  & 206 & 34.5 & 179 & 38.0 & \textbf{+3.5}  & \textbf{+1.4}  & \textbf{+5.8}  & \(\mathbf{.001}\)   & 126 & 45.2 & \textbf{+10.8} & \textbf{+5.9} & \textbf{+16.0} & \(\mathbf{< .001}\) \\
\midrule
\multirow{2}{*}{Combined} 
& GPT-4o   & 706 & 51.7 & 499 & 62.9 & \textbf{+14.2} & \textbf{+11.7} & \textbf{+16.8} & \(\mathbf{< .001}\) & 334 & 76.3 & \textbf{+24.6} & \textbf{+20.8} & \textbf{+28.5} & \(\mathbf{< .001}\) \\
& GPT-4.1  & 706 & 54.8 & 626 & 60.4 & \textbf{+5.6}  & \textbf{+4.2}  & \textbf{+7.1}  & \(\mathbf{< .001}\) & 499 & 63.8 & \textbf{+13.5} & \textbf{+11.0} & \textbf{+16.2} & \(\mathbf{< .001}\) \\
\bottomrule
\end{tabular}%
}
\end{table*}

\begin{figure}[htbp]
  \centering
  \includegraphics[width=\textwidth]{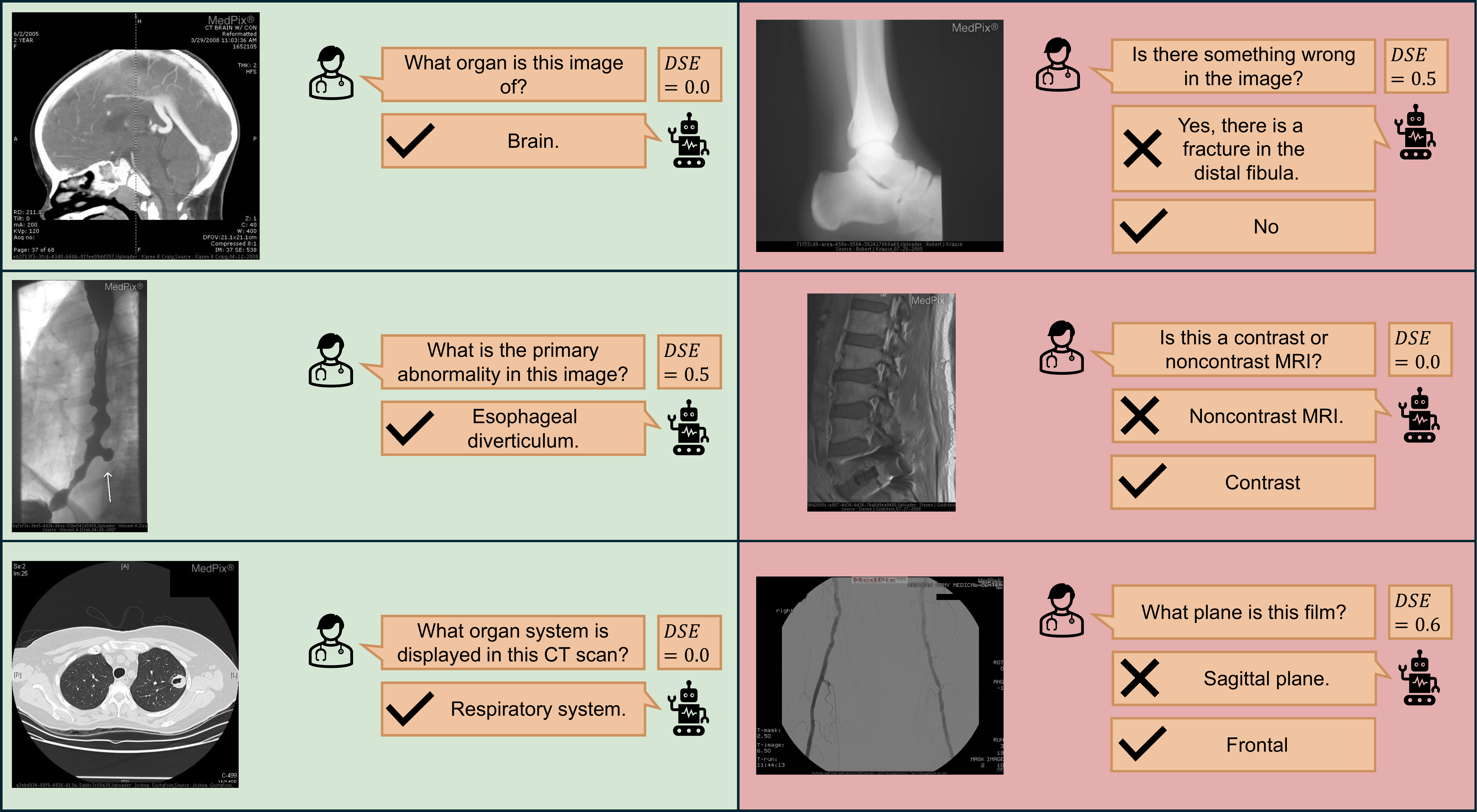}
  \caption{Examples where GPT-4o provides correct answers (left) and incorrect answers (right) on the VQA-Med 2019 dataset. If a discrete semantic entropy (DSE) threshold of 0.3 had been applied, the middle left question (DSE = 0.5) and the top and bottom questions on the right would have been filtered out. The middle question on the right illustrates a case where DSE fails to identify a hallucinated response, as the model confidently provided an incorrect answer (``Noncontrast MRI'').}\label{fig:trueFalseExample}
\end{figure}

Representative examples of correct answers, high-entropy rejections, and a case where the VLM confidently stated a hallucination are shown in Figure \ref{fig:trueFalseExample}.

Across subcategories of VQA-Med 2019, the effects of DSE-based filtering varied (Table \ref{tab:subcategory-accuracy}). At DSE $\leq$ 0.6, accuracy slightly improved for organ questions, from 69.6\% to 71.6\% for GPT-4o and minimally from 80.8\% to 81.1\% for GPT-4.1. For plane questions, the impact of DSE-based filtering was negligible, with accuracy changing from 71.2\% to 71.8\% for GPT-4o and remaining unchanged at 76.0\% for GPT-4.1. Performance on the challenging abnormality questions remained modest at this threshold (GPT-4o: 33.3\%, GPT-4.1: 16.0\%), reflecting the persistent difficulty of this category. On the RadDataset, DSE filtering at 0.3 increased GPT-4.1 accuracy on CT questions from 31.7\% at baseline to 45.2\%, with similar trends observed for other modalities. 

\begin{table*}[t]
\centering
\caption{Accuracy (Acc.) of the GPT-4o and GPT-4.1 model according to dataset subcategory. Baseline accuracy and accuracy after applying the indicated discrete semantic entropy (DSE) threshold are shown along with the number (\#) of answered questions.}
\label{tab:subcategory-accuracy}
\resizebox{\textwidth}{!}{%
\begin{tabular}{lllcccccc}
\toprule
 &  &  & \multicolumn{2}{c}{All} & \multicolumn{2}{c}{DSE $\le 0.6$} & \multicolumn{2}{c}{DSE $\le 0.3$} \\
\cmidrule(lr){4-5} \cmidrule(lr){6-7} \cmidrule(lr){8-9}
Dataset & Sub-Category & Model & \# & Acc.\ in \% & \# & Acc.\ in \% & \# & Acc.\ in \% \\
\midrule
\multirow{8}{*}{\shortstack{VQA\text{-}Med\\2019}}
& \multirow{2}{*}{Modality}    & GPT-4o   & 125 & 81.6 & 125 & 81.6 & 114 & 84.2 \\
&                                 & GPT-4.1  & 125 & 83.2 & 125 & 83.2 & 115 & 85.2 \\
\cmidrule(lr){2-9}
& \multirow{2}{*}{Plane}       & GPT-4o   & 125 & 71.2 & 124 & 71.8 & 105 & 79.0 \\
&                                 & GPT-4.1  & 125 & 76.0 & 125 & 76.0 & 117 & 77.8 \\
\cmidrule(lr){2-9}
& \multirow{2}{*}{Organ}       & GPT-4o   & 125 & 69.6 & 116 & 71.6 &  75 & 77.3 \\
&                                 & GPT-4.1  & 125 & 80.8 & 122 & 81.1 & 110 & 83.6 \\
\cmidrule(lr){2-9}
& \multirow{2}{*}{Abnormality} & GPT-4o   & 125 & 13.6 &  27 & 33.3 &  11 & 36.4 \\
&                                 & GPT-4.1  & 125 & 12.8 &  75 & 16.0 &  31 &  9.7 \\
\midrule
\multirow{8}{*}{RadDataset}
& \multirow{2}{*}{Angiography} & GPT-4o   & 26 & 50.0 & 16 & 62.5 &  8 & 75.0 \\
&                                 & GPT-4.1  & 26 & 50.0 & 25 & 52.0 & 21 & 61.9 \\
\cmidrule(lr){2-9}
& \multirow{2}{*}{CT}          & GPT-4o   & 60 & 35.0 & 29 & 37.9 &  5 & 20.0 \\
&                                 & GPT-4.1  & 60 & 31.7 & 52 & 34.6 & 31 & 45.2 \\
\cmidrule(lr){2-9}
& \multirow{2}{*}{MRI}         & GPT-4o   & 60 & 28.3 & 34 & 38.2 & 10 & 30.0 \\
&                                 & GPT-4.1  & 60 & 23.3 & 54 & 25.9 & 42 & 28.6 \\
\cmidrule(lr){2-9}
& \multirow{2}{*}{Radiography} & GPT-4o   & 60 & 31.7 & 28 & 42.9 &  6 & 66.7 \\
&                                 & GPT-4.1  & 60 & 41.7 & 48 & 47.9 & 32 & 56.2 \\
\bottomrule
\end{tabular}%
}
\end{table*}

\begin{figure}[htbp]
  \centering
  \includegraphics[width=\textwidth]{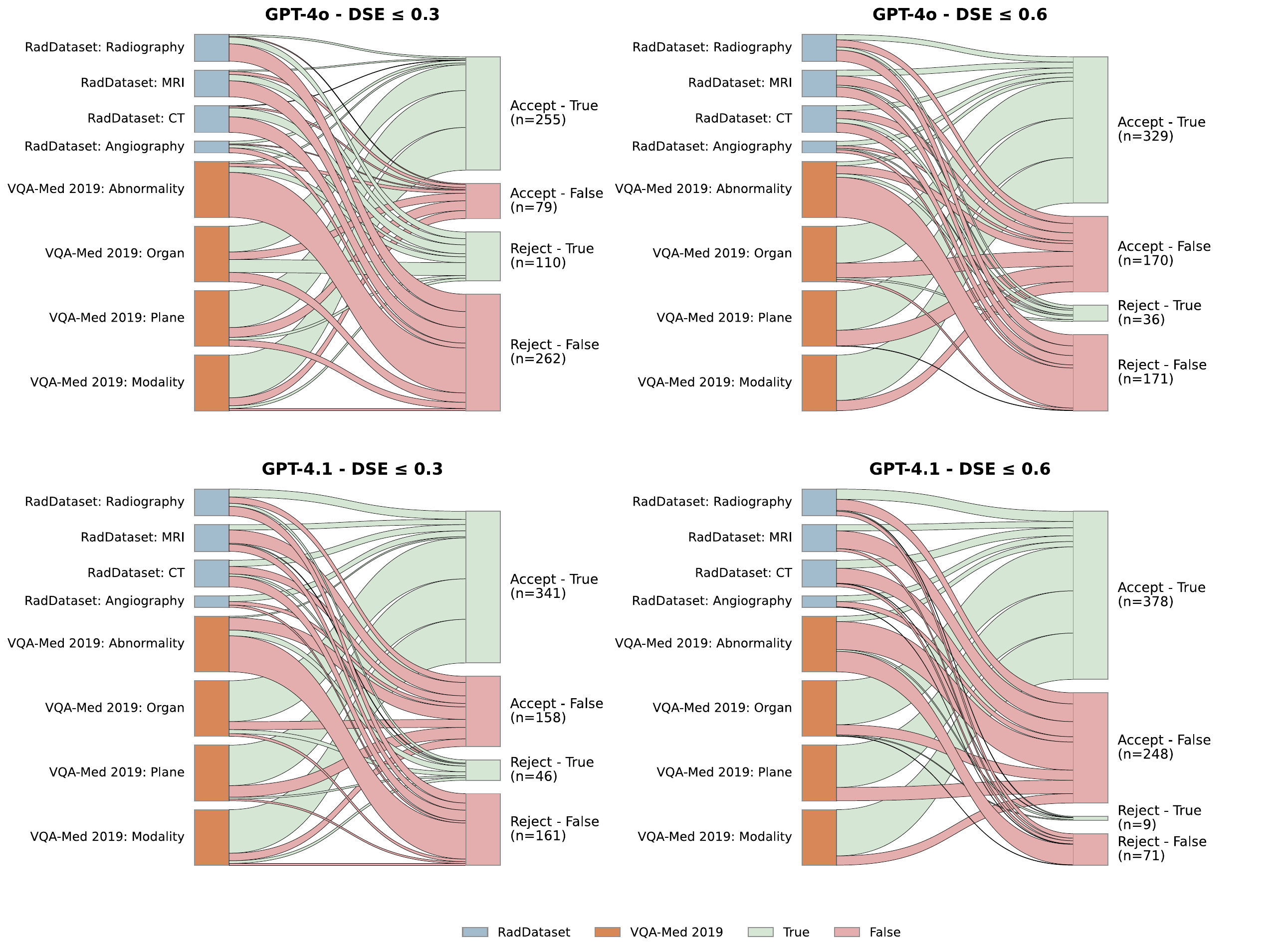}
  \caption{Sankey diagrams depicting the distribution of accepted and rejected answers across vision–language models (GPT-4o, GPT-4.1) and discrete semantic entropy (DSE) thresholds ($\leq$ 0.6 and $\leq$ 0.3). Each flow originates from question subcategories within the VQA-Med 2019 and RadDataset cohorts and terminates in one of four outcome groups: accepted true, accepted false, rejected true, or rejected false. The proportion of true versus false answers varies substantially across subcategories. Tightening the DSE threshold from 0.6 to 0.3 increases the number of rejected questions and correspondingly improves the accuracy of retained answers by filtering semantically inconsistent responses.}\label{fig:sankex2x2}
\end{figure}

The number of rejected questions showed large differences in the subcategories (Table \ref{tab:subcategory-accuracy},Figure \ref{fig:sankex2x2}). In the modality prediction subcategory of the VQA-Med 2019 dataset, both GPT-4o and GPT-4.1 demonstrated high baseline accuracy (81.6\% and 83.2\%, respectively). Applying DSE filtering at a threshold of 0.6 resulted in no rejected questions and no accuracy change for either model. Even at the stricter DSE threshold of $\leq$ 0.3, rejection rates remained low, with only 8.8\% of questions removed (GPT-4o: 11/125, GPT-4.1: 8.0\%, 10/125), leading to marginal accuracy improvements of +2.6\% and +2.0\%, respectively.

In contrast, the imaging plane and organ subcategories showed moderate rejection rates when applying a DSE $\leq$ 0.3 threshold (GPT-4o: 16.0\% plane, 40.0\% organ), with corresponding accuracy gains of +7.8\% and +7.7\%. The abnormality subcategory exhibited notably higher rejection rates, reaching 78.4\% at DSE $\leq$ 0.6 for GPT-4o and exceeding 91.2\% at DSE $\leq$ 0.3. GPT-4.1 showed lower rejection at the same thresholds (40.0\% at 0.6) but less consistent accuracy improvements. GPT-4o achieved the highest observed accuracy increase in this subcategory (+19.7\%), while GPT-4.1 displayed a smaller and variable effect (+3.2\% at 0.6, -3.1\% at 0.3).

\begin{figure}[htbp]
  \centering
  \includegraphics[width=\textwidth]{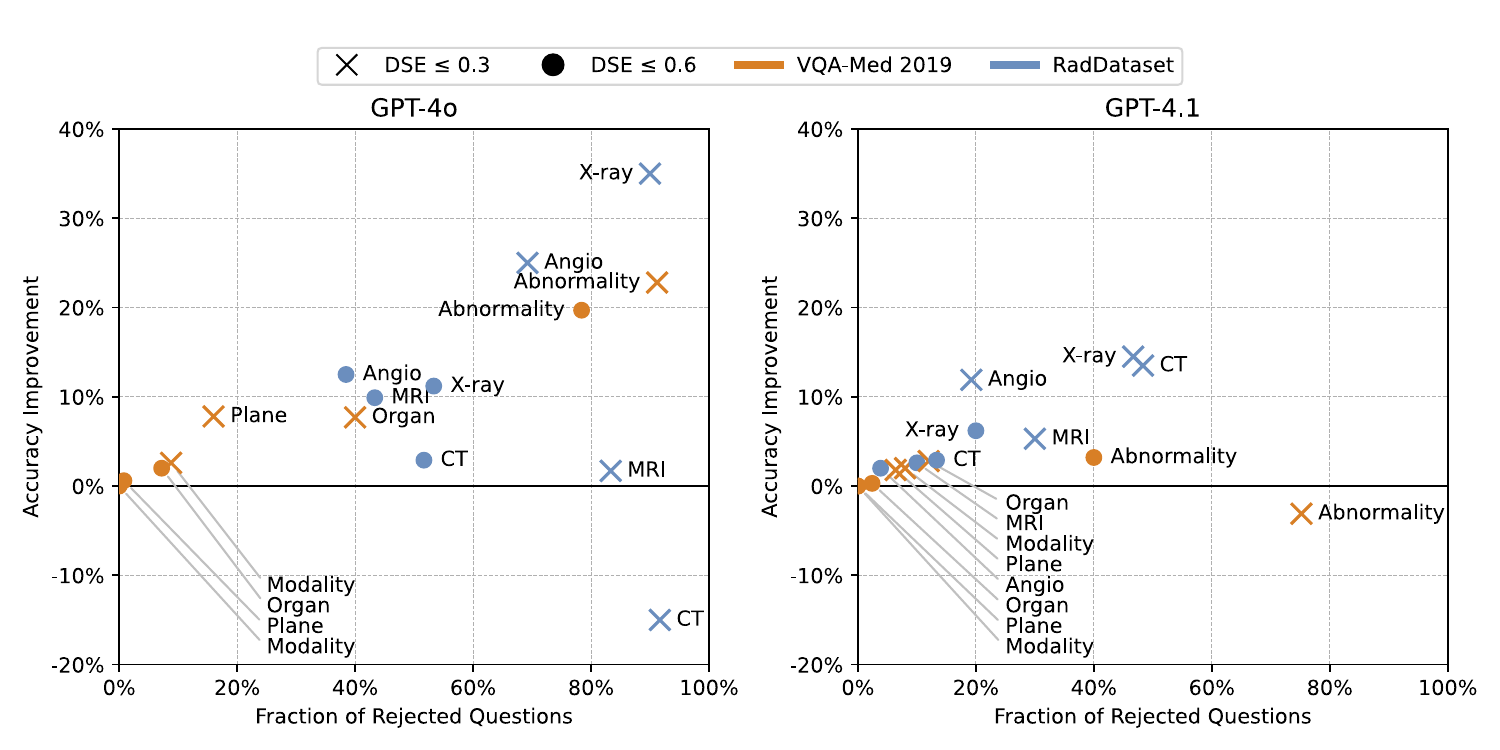}
  \caption{Relative accuracy improvement versus reduction in the number of answered questions for GPT-4o (left) and GPT-4.1 (right) after applying discrete semantic entropy (DSE) thresholds of 0.6 and 0.3. Each point represents a subcategory within the evaluated datasets (e.g., organ, abnormality, modality, imaging plane). Stricter thresholds (DSE $\leq$ 0.3) generally yield higher accuracy improvements at the cost of answering fewer questions, illustrating the trade-off between reliability and coverage across subcategories.}\label{fig:relativeSubCathegoryImprovement}
\end{figure}

On the RadDataset, modality-specific differences were evident. Angiography questions saw high rejection rates at DSE $\leq$ 0.3 (GPT-4o: 69.2\%, GPT-4.1: 19.2\%) and corresponding accuracy increases to 75.0\% and 61.9\%, respectively. In contrast, CT and MRI filtering was extensive, particularly for GPT-4o, which retained only 8.3\% of CT questions and experienced a decline in accuracy for this modality (-15.0\%). Figure \ref{fig:relativeSubCathegoryImprovement} presents a comparison of model performance and rejection rates across subcategories.

\begin{figure}[htbp]
  \centering
  \includegraphics[width=\textwidth]{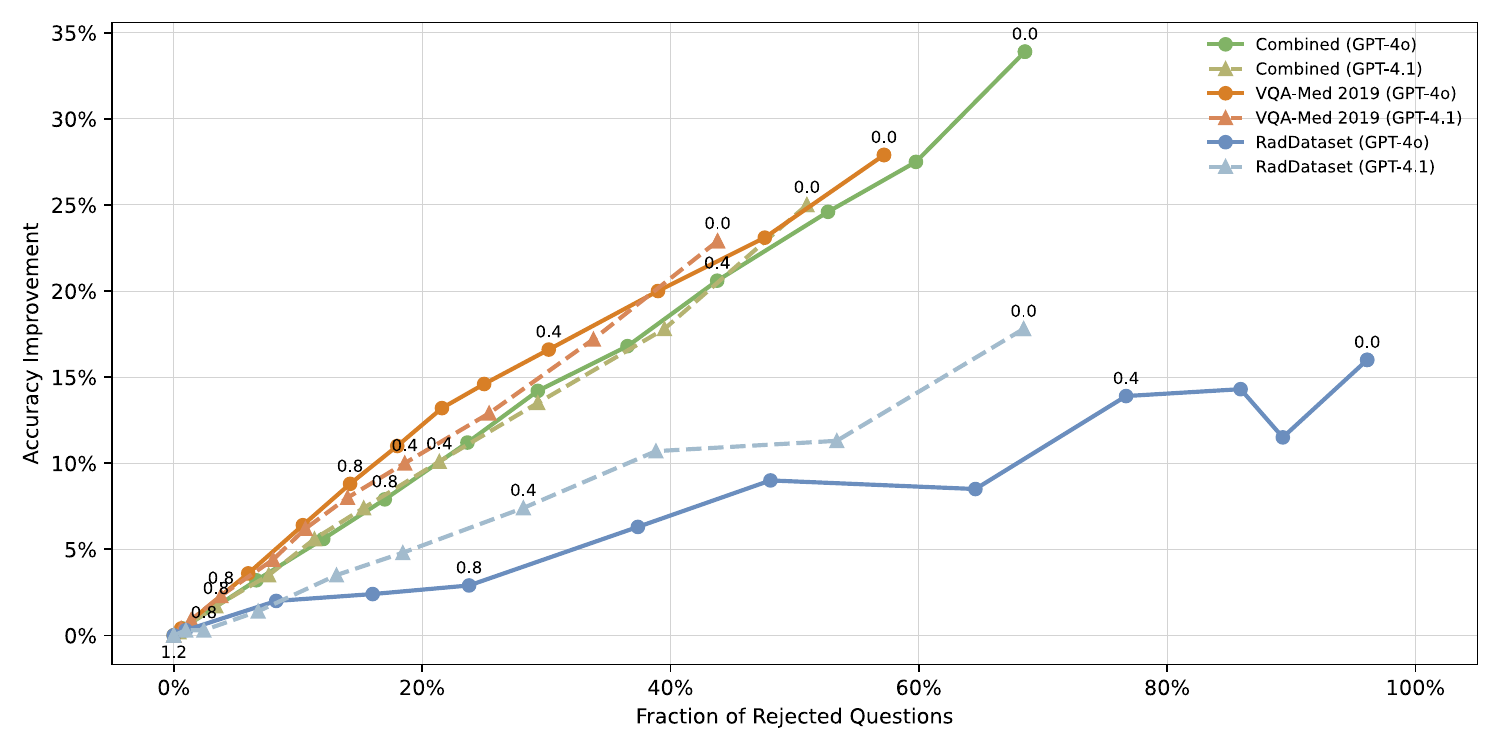}
  \caption{Accuracy–coverage trade-off under discrete semantic entropy (DSE) filtering. Curves show the absolute accuracy gain (y-axis) versus the fraction of rejected questions (x-axis) as the DSE threshold (numeric labels next to points) is varied. For each setting, questions with DSE above the threshold are discarded. Lower thresholds (stricter filtering) reject more questions but yield larger accuracy improvements. A threshold of 1.2 corresponds to no filtering. }\label{fig:rejectionImprovement}
\end{figure}

As visualized in Figure \ref{fig:rejectionImprovement}, reducing the DSE threshold increased the accuracy of the responses to the remaining questions but reduced the number of questions answered. For example, at DSE $\leq$ 0.3, only 334 of 706 (47.3\%) questions were answered by GPT-4o, but with an accuracy gain of +24.6\%. Figure \ref{fig:rejectionImprovement} illustrates this trade-off: as the DSE threshold decreases, accuracy rises, but the number of rejected questions as well. 

\subsection*{Implementation Feasibility, Latency, and Cost}
The DSE workflow was designed for practical implementation, relying exclusively on standard API calls without needing access to model internals like token probabilities, ensuring identical application across different VLMs. In terms of latency, the pipeline architecture permits substantial parallelization. The two primary computational steps—the generation of $k$ sampled answers and the subsequent $k(k-1)$ pairwise entailment checks—can each be performed concurrently. Consequently, the total pipeline latency is approximately twice that of a single API call. For the GPT-4o model, we measured a mean single-call latency of approximately 3 seconds (standard deviation: $\sim 2$ s), resulting in a total execution time of $\sim6$ seconds for the fully parallelized DSE pipeline.
From a cost perspective, our analysis was based on an average price of \$10 per one million tokens for the GPT-4o model. The total cost to process one image-question pair broke down into the answer sampling ($\sim$\$0.11), and the subsequent entailment checks ($\sim$\$0.61). This culminated in a total cost of approximately \$0.72 per question, an amount that suggests the method is financially feasible for integration into daily clinical workflows.

\section{Discussion}
The integration of advanced AI, particularly VLMs, into radiology holds promise for alleviating workload pressures and supporting diagnostic tasks\cite{european2022roleS2,bhayana2024chatbotsS7}. However, the tendency of these models to generate plausible but factually incorrect outputs, or ``hallucinations,'' presents a significant barrier to safe clinical adoption\cite{Gunjal2024DetectionS13,farquhar2024detectingS14}. This study investigated DSE—a measure of semantic consistency across multiple model responses—as a practical method for identifying and filtering unreliable VLM outputs in radiologic image interpretation tasks, extending DSE from its original text-only application\cite{farquhar2024detectingS14} to the multimodal domain of radiology. Our central finding is that applying DSE filtering significantly improves the accuracy of accepted VLM responses by selectively rejecting questions associated with high semantic inconsistency, thereby enhancing model trustworthiness. 

A key advantage of DSE is its applicability to black-box models. Unlike methods that require access to internal model states such as probability distributions\cite{Ji2023MAPS16,ZOU2023100003S17} or need additional training data or fine-tuning for auxiliary components like uncertainty classifiers\cite{Gunjal2024DetectionS13}, DSE operates solely on the model's outputs generated via standard application programming interface (API) calls. This makes it readily deployable alongside proprietary VLMs without requiring specific cooperation from model developers. It provides a practical alternative or complement to other black-box consistency checks, such as those based on input paraphrasing\cite{Khan2024BlackBoxS19} or multi-report sampling\cite{pmlr-v259-zhang25cS18}, by focusing directly on the semantic stability of the model's own generated answers to a single, consistent query. Compared with RadFlag\cite{pmlr-v259-zhang25cS18}, question rephrasing\cite{Khan2024BlackBoxS19}, or vision-amplified semantic entropy\cite{Liao2025VisionAplifiedS20}, DSE avoids input modifications or long multi-report generation, offering lower latency and broader applicability. In practice, DSE incurs substantially lower token and computational cost than multi-report approaches such as RadFlag or paraphrase-consistency methods, because it reuses a single prompt and short answers instead of repeatedly rephrasing inputs or generating full-length reports. While RadFlag excels for full-report tasks but is computationally heavy, and rephrasing or vision-amplified methods risk altering clinical nuance or image fidelity, DSE focuses on semantic consistency alone—efficient yet still limited when hallucinations are repeated with high confidence.

Integrating DSE into PACS or reporting systems can streamline workflows by screening queries at the question level to either withhold unstable outputs or attach an interpretable uncertainty score derived from semantic dispersion across sampled responses. In practice, this identifies questions that are more likely to elicit hallucinations and, when answers are returned, provides a quantitative signal that radiologists can reference when judging the safety of acting on a given question–answer pair. Making uncertainty explicit is essential in clinical settings, where decisions depend on calibrated confidence rather than linguistic fluency. Because DSE operates solely on model outputs and requires neither access to token probabilities nor model retraining, it can be deployed as a lightweight wrapper around existing VLM integrations. By combining selective answering with transparent uncertainty, DSE can thus strengthen clinician trust and real-world acceptance of vision–language models in radiology. In clinical practice, selecting a DSE threshold requires balancing coverage against acceptable risk. For applications such as diagnostic decision support, we recommend a stricter threshold (e.g., 0.3 in our experiments) to prioritize specificity, minimizing the risk of incorrect advice by aggressively rejecting uncertain predictions. Conversely, in lower-risk scenarios like human-in-the-loop screening, a more lenient threshold (e.g., 0.6) may be preferable to prioritize sensitivity, ensuring that potentially relevant information is not discarded. These specific values are model-dependent, and the exact operating point should be determined via calibration on a held-out dataset for each clinical application.

Despite its promise, DSE has important limitations. First, it measures semantic consistency, not factual correctness\cite{farquhar2024detectingS14}. A VLM could consistently generate the same incorrect answer (a "confident hallucination"), resulting in a low DSE score that would bypass the filter. This represents a significant residual risk. An example of this is illustrated by the middle right panel of Figure \ref{fig:trueFalseExample}. Integrating DSE with complementary uncertainty signals (e.g., linguistic uncertainty cues, model calibration techniques if available) could lead to more robust hallucination detection systems. Accordingly, DSE should be interpreted as a selective-answering uncertainty signal rather than a guarantee of correctness; in safety-critical use cases, outputs should not be acted upon without radiologist verification. While additional validated clinical context or cross-model consistency checks may help, they are unlikely to eliminate this failure mode; approaching near-zero risk would require restricting the system to a closed set of vetted questions/answers (e.g., guideline- or database-backed retrieval), shifting from open-ended generation to curated decision support. Second, the reliability of the semantic clustering step relies on the VLM's own entailment capabilities. Errors or biases in the VLM's understanding of semantic equivalence, particularly for nuanced or complex clinical statements, could lead to inaccurate DSE calculations. This limitation could be mitigated by using an external, clinically validated entailment model for clustering, reducing dependence on the VLM under evaluation.  Third, while evaluated on two datasets including clinical diagnostic questions, further validation is essential across a broader range of imaging modalities, pathologies, patient populations, and clinical query types to establish generalizability. Fourth, our evaluation was restricted to 2D images. VQA-Med 2019 consists of single 2D images, and in the RadDataset we used manually selected key slices extracted from volumetric CT, MR, radiography, and angiography examinations. This design assumes that the most informative slice has already been identified and therefore does not fully capture the challenges of applying DSE to full 3D studies in routine practice. As a result, our findings should be interpreted as a best-case estimate for question-level hallucination filtering under simplified viewing conditions. Extending DSE-based uncertainty filtering to volumetric workflows, for example by integrating it with automatic key-slice selection or with VLMs that natively operate on 3D image stacks, is an important direction for future work. Prospective clinical validation studies are also needed to evaluate the real-world impact of DSE-filtered VLM outputs on radiologist workflow, diagnostic confidence, efficiency, and, ultimately, patient outcomes. Such studies would also provide insights into how radiologists interact with and trust AI systems employing this type of uncertainty filtering. Whether high rejection rates increase clinician trust or render the system impractical, and whether selective answering might inadvertently promote over-reliance on the remaining outputs despite limited accuracy, are open questions that require prospective user studies to resolve. Finally, a stricter DSE threshold yielded higher accuracy gains at the cost of answering fewer questions, highlighting the inherent trade-off between accuracy and answer rate (coverage). The optimal DSE threshold will likely depend on the specific clinical application and the tolerance for potentially incorrect AI suggestions versus the need for comprehensive assistance. We also observed that the two evaluated OpenAI models differed in how rapidly accuracy improved under stricter numeric DSE thresholds (Fig. 6), reflecting that models can exhibit different degrees of answer variability (and thus different DSE distributions) even when queried with the same temperature. Because the model internals and decoding details are not fully transparent, we cannot further analyze or technically attribute this effect; however, it is an important practical consideration when transferring DSE-based filtering to additional models, as thresholds should be calibrated per model. Prior work on semantic entropy has demonstrated consistent effects across models from different providers\cite{farquhar2024detectingS14}, suggesting that DSE-based filtering is likely transferable to other VLMs. The observed performance differences between datasets and VQA subcategories underscore the sensitivity of both VLMs and DSE filtering to the specific task context. 

In conclusion, discrete semantic entropy offers a practical method to quantify uncertainty and filter unreliable outputs from black-box VLMs applied to radiologic image interpretation. The VLMs generally exhibited low baseline accuracy on complex interpretive tasks, especially for abnormality detection, reinforcing that current general-purpose VLMs are not yet suitable for autonomous diagnostic interpretation. This work should therefore be viewed as a proof-of-concept demonstration of applying DSE to radiologic data, not as establishing clinically practical operating points, especially for abnormality detection where rejection rates are high. By enabling selective prediction based on semantic consistency, DSE significantly improves the percentage of accurate responses, a step towards enhancing the safety and trustworthiness of AI tools in radiology. While DSE is not a panacea for the current limitations of VLMs, particularly in complex diagnostic tasks, it provides a crucial mechanism for managing uncertainty as these powerful models are increasingly explored for clinical support roles in radiology practice.

\backmatter

\bmhead{Acknowledgements}
This research is supported by the Deutsche Forschungsgemeinschaft - DFG (NE 2136/7-1, NE 2136/3-1, TR 1700/7-1), the German Federal Ministry of Research, Technology and Space (Transform Liver - 031L0312C, DECIPHER-M - 01KD2420B) and the European Union Research and Innovation Programme (ODELIA - GA 101057091). The authors gratefully acknowledge the computing time provided to them at the NHR Center NHR4CES at RWTH Aachen University (project number p0021834). This is funded by the Federal Ministry of Education and Research, and the state governments participating on the basis of the resolutions of the GWK for national high performance computing at universities (www.nhr-verein.de/unsere-partner). The data used in this publication were managed using the research data management platform Coscine (http://doi.org/10.17616/R31NJNJZ) with storage space of the Research Data Storage (RDS) (DFG: INST222/1261-1) and DataStorage.nrw (DFG: INST222/1530-1) granted by the DFG and Ministry of Culture and Science of the State of North Rhine-Westphalia. J.N.K. is supported by the German Cancer Aid DKH (DECADE, 70115166), the German Federal Ministry of Research, Technology and Space BMFTR (PEARL, 01KD2104C; CAMINO, 01EO2101; TRANSFORM LIVER, 031L0312A; TANGERINE, 01KT2302 through ERA-NET Transcan; Come2Data, 16DKZ2044A; DEEP-HCC, 031L0315A; DECIPHER-M, 01KD2420A; NextBIG, 01ZU2402A), the German Research Foundation DFG (CRC/TR 412, 535081457; SFB 1709/1 2025, 533056198), the German Academic Exchange Service DAAD (SECAI, 57616814), the German Federal Joint Committee G-BA (TransplantKI, 01VSF21048), the European Union EU’s Horizon Europe research and innovation programme (ODELIA, 101057091; GENIAL, 101096312), the European Research Council ERC (NADIR, 101114631), the National Institutes of Health NIH (EPICO, R01 CA263318) and the National Institute for Health and Care Research NIHR (Leeds Biomedical Research Centre, NIHR203331). The views expressed are those of the author(s) and not necessarily those of the NHS, the NIHR or the Department of Health and Social Care. This work was funded by the European Union. Views and opinions expressed are, however, those of the author(s) only and do not necessarily reflect those of the European Union. Neither the European Union nor the granting authority can be held responsible for them. We used generative AI tools for language editing and rephrasing; all scientific content, data, analyses, and conclusions were written and verified by the authors.

\bmhead{Competing interests}
D.T. received honoraria for lectures by Bayer, GE, Roche, AstraZenica, and Philips and holds shares in StratifAI GmbH, Germany and in Synagen GmbH, Germany. J.N.K. declares consulting services for AstraZeneca, MultiplexDx, Panakeia, Mindpeak, Owkin, DoMore Diagnostics, and Bioptimus. Furthermore, he holds shares in StratifAI, Synagen, and Spira Labs, has received an institutional research grant from GSK, and has received honoraria from AstraZeneca, Bayer, Daiichi Sankyo, Eisai, Janssen, Merck, MSD, BMS, Roche, Pfizer, and Fresenius.

\bibliography{bibliography}

\end{document}